%% file: od_prediction.tex
\documentclass[11 pt, onecolumn]{article}

\usepackage[margin=1.1in]{geometry}
\input{./sections/header}
\usepackage{algorithm}
\usepackage{algpseudocode}
\usepackage{caption}

\DeclareMathAlphabet{\mathcal}{OMS}{cmsy}{m}{n}

\usepackage{graphicx}
\usepackage{epstopdf}

\begin{document}
\title{\LARGE \bf Toward Automated Regulatory Decision-Making: Trustworthy Medical Device Risk Classification with Multimodal Transformers and Self-Training}

\author{Yu Han, Aaron Ceross, and Jeroen H.M. Bergmann
\thanks{Yu Han is with the Institute of Biomedical Engineering, Department of Engineering Science, University of Oxford, Oxford, UK. Aaron Ceross is with Birmingham Law School, University of Birmingham, Birmingham, UK. Jeroen H.M. Bergmann is with the Institute of Biomedical Engineering, University of Oxford, and the Department of Technology and Innovation, University of Southern Denmark, Odense, Denmark, emails: yu.han@ox.ac.uk, a.ceross@bham.ac.uk, jeroen.bergmann@eng.ox.ac.uk}%
}
\newcommand*{\QEDA}{\hfill\ensuremath{\blacksquare}}%

\maketitle

\begin{abstract}
Accurate classification of medical device risk levels is essential for regulatory oversight and clinical safety. We present a Transformer-based multimodal framework that integrates textual descriptions and visual information to predict device regulatory classification. The model incorporates a cross-attention mechanism to capture intermodal dependencies and employs a self-training strategy for improved generalization under limited supervision. Experiments on a real-world regulatory dataset demonstrate that our approach achieves up to 90.4\% accuracy and 97.9\% AUROC, significantly outperforming text-only (77.2\%) and image-only (54.8\%) baselines. Compared to standard multimodal fusion, the self-training mechanism improved SVM performance by 3.3 percentage points in accuracy (from 87.1\% to 90.4\%) and 1.4 points in macro-F1, suggesting that pseudo-labeling can effectively enhance generalization under limited supervision. Ablation studies further confirm the complementary benefits of both cross-modal attention and self-training. We also evaluate the model’s robustness to noisy inputs and modality-specific perturbations, demonstrating resilience in low-resource settings. Beyond technical gains, we highlight applications in automated pre-screening, compliance verification, and support for harmonization frameworks such as UDI and GMDN. Our findings suggest that domain-adapted multimodal models, when designed with transparency and specificity, can serve as trustworthy decision-support tools for regulatory classification tasks.
\end{abstract}


\input{./sections/introduction} 
\input{./sections/model} 
\input{./sections/experiments} 
\input{./sections/conclusion} 

\bibliographystyle{IEEEtran}
\bibliography{od_prediction}   
\end{document}

%% file: sections/header.tex
\usepackage{amsmath}
\usepackage{bm}	
\usepackage{graphicx}

\usepackage{graphicx}
\usepackage{multirow}
\usepackage{mdwlist}

\usepackage{threeparttable}

\usepackage{amsfonts}
\usepackage{amsmath}
\usepackage{hyperref}
\usepackage{bbm}
\usepackage{float}
\usepackage{caption}
\usepackage{subcaption}
\usepackage{xcolor}
\usepackage{mathtools}
\usepackage{tikz}

\usetikzlibrary{shapes.geometric,arrows}
\usepackage{enumerate}
\usepackage{multicol}
\usepackage{stackrel}

\usepackage{subcaption}
\usepackage{amssymb}
\usepackage[utf8]{inputenc}
\usepackage{booktabs,lipsum}
\usepackage{hyperref}

%% file: sections/introduction.tex
\section{Introduction}
Medical device classification is a foundational task in regulatory compliance, inventory management, and clinical decision-making. Regulatory authorities such as the U.S. Food and Drug Administration (FDA) and China's National Medical Products Administration (NMPA) assign risk categories (Class I, II, III) to devices based on their intended use, invasiveness, and criticality \cite{han2024regulatory}. These classifications inform market approval pathways, post-market surveillance, and the stringency of clinical evidence requirements. Traditionally, classification has relied heavily on manual expert review of device descriptions and product images—a process that is labor-intensive, error-prone, and susceptible to inter-rater variability. The implications of misclassification are profound. Under-classified devices may bypass essential safety evaluations, posing risks to patients and liability challenges to manufacturers. Over-classified devices, conversely, may face unnecessary regulatory burdens, delaying access to beneficial technologies and stifling innovation. This problem affects a wide range of stakeholders, including regulatory reviewers, industry sponsors, healthcare providers, and, ultimately, patients. As the volume and complexity of medical device submissions increase—particularly with the rise of digital health products, AI-based software, and cross-border regulation—the need for scalable, robust, and interpretable classification support tools has become increasingly urgent \cite{han2024more}.

Within the broader context of regulatory affairs, medical device classification serves as a representative use case for several reasons. First, it involves decision-making based on heterogeneous, often unstructured information sources—ranging from technical specifications and clinical indications to product images and labeling. This reflects a common pattern across regulatory tasks, where judgments must be rendered from a mix of textual, visual, and structured inputs. Second, classification interacts with other core regulatory processes, including labeling verification, predicate device matching, conformity assessment, and regulatory pathway selection. Improvements in classification, therefore, have downstream benefits across the regulatory pipeline.

In recent years, multimodal machine learning has emerged as a powerful paradigm for integrating heterogeneous data sources—including text, images, audio, and video into unified representations for downstream tasks~\cite{baltrusaitis2019multimodal, poria2017review, chen2020simple}. In domains such as e-commerce~\cite{baltrusaitis2019multimodal}, social media~\cite{poria2017review}, and autonomous driving~\cite{chen2020simple}, joint modeling of visual and textual modalities has demonstrated significant gains in classification, retrieval, and reasoning. In healthcare, the availability of narrative descriptions and visual content (e.g., radiographs, clinical photographs) makes multimodal modeling particularly promising for diagnostic support~\cite{jing2018automatic}, report generation~\cite{liu2021aligntransformer}, and biomedical information extraction~\cite{lu2009biocreative}.

Despite the growing success of multimodal learning in other domains, its application to regulatory classification of medical devices remains limited. Current computational approaches in regulatory affairs field are still nascent \cite{han2024evaluation}. While regulatory records often contain both structured fields (e.g., registration numbers) and unstructured modalities such as textual descriptions and product images, most existing models rely solely on text-based inputs~\cite{zhang2021textclass, ceross2021machine}. This unimodal focus overlooks complementary visual information that may be essential for accurate risk stratification. The limitation becomes especially pronounced in cases where textual descriptions are ambiguous, incomplete, or heavily reliant on domain-specific terminology that may obscure critical distinctions.

To address these limitations, we propose a novel deep learning framework that integrates textual and visual data using a cross-attention Transformer architecture. Our system learns joint representations that capture fine-grained interdependencies across modalities, enabling it to reconcile semantic ambiguity and leverage complementary signals. Inspired by recent advances in vision-language modeling~\cite{li2019visualbert, tan2019lxmert}, we also incorporate a self-training mechanism~\cite{xie2020self} to expand the labeled dataset using high-confidence pseudo-labels, thereby improving model robustness under limited supervision.

Beyond methodological motivations, the integration of visual data in medical device classification is also driven by practical and industrial considerations. In real-world manufacturing and regulatory workflows, visual inspection of devices is an essential step for verifying labeling consistency, detecting counterfeit products, and ensuring compliance with international standards such as the Global Medical Device Nomenclature (GMDN) or the Unique Device Identification (UDI) system. Manufacturers frequently maintain internal product catalogs that include both text records and standardized imagery, making image-based classification a natural complement to traditional document review. For example, in large-scale factories, barcode-scanned images of devices can be automatically matched against regulatory labels to validate their assigned class. In cross-border scenarios, where documentation may be incomplete or formatted differently due to language and jurisdictional differences, visual features provide a universal modality to anchor classification decisions. Furthermore, visual models may assist in real-time inventory auditing, post-market surveillance, or customs clearance by identifying product types through photographic cues, especially in fast-moving supply chains.

Importantly, as computer vision technologies continue to advance—driven by progress in transformers, contrastive learning, and foundation models—the potential for high-fidelity image understanding in medical settings will only increase. Future applications could include automated comparison of devices with prototype schematics, identification of unregistered knockoffs through visual anomaly detection, or the reconstruction of regulatory attributes (e.g., material composition, interface type) from surface-level appearance. In this context, learning joint representations from both textual and visual information not only enhances classification accuracy but also lays the groundwork for more intelligent, scalable, and globally interoperable regulatory systems.

Our main contributions are four-fold:
\begin{itemize}
    \item We construct a hybrid textual feature extraction pipeline that combines TF-IDF, latent semantic analysis (LSA), and Sentence-BERT to capture both shallow and contextual semantics.
    \item We extract domain-adapted image representations using a fine-tuned EfficientNet-B4 architecture, enhanced with attention modules for improved feature salience.
    \item We introduce a novel cross-modal Transformer encoder that performs fusion via stacked self-attention layers, enabling the model to learn from contextual interactions between modalities.
    \item We implement a multi-round pseudo-label self-training strategy that augments the training data with high-confidence predictions, improving generalization without additional annotations.
\end{itemize}

\section{Related Work}

Multimodal learning has become an increasingly important paradigm for integrating heterogeneous inputs—such as text, images, and structured meta data into unified models for classification, retrieval, and reasoning tasks. This section reviews representative approaches to multimodal fusion, with a focus on their applicability to healthcare and regulatory scenarios. We also highlight how our proposed method differs from general-purpose vision-language models such as CLIP, VisualBERT, and LXMERT.

\subsection*{Multimodal Fusion Architectures}

Early fusion techniques merge feature vectors from each modality (e.g., TF-IDF and CNN outputs) before training a joint classifier~\cite{bruni2011distributional, kiela2014learning}. While this approach is simple and effective for aligned inputs, it lacks the capacity to model inter-modal dependencies and is sensitive to missing or noisy modalities --- issues that are common in real-world regulatory data. Late fusion methods instead maintain separate pipelines for each modality and combine their outputs at the decision level using ensembling strategies such as voting or stacking~\cite{rokach2010ensemble}. These methods are modular and robust to modality imbalance, but they are limited in capturing the nuanced interactions that may exist between, for example, a textual description of use-case and the visual complexity of a medical device.

To overcome these limitations, attention-based models and cross-modal Transformers have been proposed~\cite{vaswani2017attention}. Models like VisualBERT~\cite{li2019visualbert}, LXMERT~\cite{tan2019lxmert}, and UNITER~\cite{chen2020simple} encode interdependencies by allowing representations from one modality to influence those of another. These architectures have achieved strong results in general-purpose vision-language benchmarks and downstream tasks such as image captioning, VQA, and retrieval.

\subsection*{Domain-Specific Models vs. Foundation Models}

General vision-language models like CLIP~\cite{radford2021learning} are trained on large-scale web image-text pairs using contrastive learning to support open-vocabulary zero-shot tasks. However, CLIP’s objective is not tailored for domain-specific classification tasks with structured taxonomies or regulatory constraints. Moreover, CLIP lacks fine-grained fusion between modalities and offers limited interpretability—critical factors in regulatory applications where trust and traceability are paramount.

In contrast, our work focuses on supervised, domain-adapted classification rather than zero-shot inference. We combine a cross-attention Transformer fusion architecture with fine-tuned text and image encoders adapted to regulatory data. Unlike CLIP, we do not treat modalities symmetrically in a contrastive setup but fuse them hierarchically to exploit their complementary structure. This enables improved handling of modality asymmetry, where either textual or visual content may dominate depending on the device type or submission format.

\subsection*{Multimodal Learning in Healthcare and Regulation}

While multimodal learning has gained traction in clinical domains—e.g., combining radiographs and reports for diagnostic prediction~\cite{huang2021fusion}, the use in regulatory contexts remains limited. Most prior work in medical device classification relies on structured text or metadata alone~\cite{zhang2021textclass, guo2022automated}, with little integration of visual content such as labeling images or product photos. Visual cues, however, often capture critical risk-related features that are not explicitly documented in text, such as device shape, ports, or intended use environment.

To address these gaps, our approach leverages cross-modal fusion tailored to the regulatory setting, enhanced by a self-training mechanism that expands the labeled dataset with high-confidence pseudo-labels. This allows the model to generalize better in low-resource conditions—common in regulatory datasets—and improve robustness to modality imbalance.

%% file: sections/model.tex
\section{Methodology}
\subsection{Problem Definition}

Let $ \mathcal{D} = \{(x_i^T, x_i^I, y_i)\}_{i=1}^N $ denote the dataset, where $ x_i^T $ and $ x_i^I $ represent the textual and image inputs respectively, and $ y_i \in \{1, 2, 3\} $ denotes the regulatory risk class.

Our goal is to learn a classifier $ f $ that predicts the correct class label based on both modalities:

$
f(x^T, x^I) \rightarrow y
$

that accurately predicts the risk class from multimodal inputs.

\subsection{Dataset and Preprocessing}
Our dataset is curated from publicly available records maintained by the National Medical Products Administration (NMPA). Each record includes a unique registration number and a product narrative description (denoted as \texttt{cpms}). Since regulatory authorities do not currently mandate the submission of product images, we supplemented the dataset by identifying and retrieving associated images from manufacturers’ official websites, linking each image to its corresponding registry entry. To establish a multimodal learning corpus, we parsed 1,000 medical device entries, matched with corresponding product images named using a unique identifier (\texttt{No-*.jpg}), and extracted structured metadata fields where available.

The target variable for classification—the device’s regulatory risk class (Class I, II, or III)—was inferred from the ninth character of the NMPA registration number (\texttt{zczbhhzbapzbh}), following the NMPA’s internal encoding scheme. Entries with missing, malformed, or ambiguous codes were excluded. We also filtered out underrepresented classes with fewer than three valid samples to ensure statistical significance during training. The final dataset includes both textual and visual modalities aligned across 1000 devices.

\subsection*{Textual Preprocessing and Semantic Embedding}

Medical device descriptions are typically sparse, domain-specific, and highly variable in both length and vocabulary. To enhance feature robustness and improve model generalization, we designed a three-stage textual feature extraction pipeline.

The raw text was first preprocessed and tokenized using Jieba, a widely used Chinese word segmentation tool \cite{sun2012jieba}. To enhance lexical diversity and simulate natural variation in expression, we applied domain-specific synonym replacement—for example, substituting terms like "device" with semantically similar alternatives such as "instrument" or "apparatus." We explored shallow semantic features by extracting term frequency-inverse document frequency (TF-IDF) vectors \cite{salton1988term} and applying Latent Semantic Analysis (LSA) \cite{deerwester1990indexing} for low-rank dimensionality reduction. This provided interpretable, coarse-grained representations of term co-occurrence patterns.

Next, we generated deep contextual embeddings using the bert-base-chinese model \cite{devlin2019bert}. Each device description was tokenized, passed through the model, and represented by averaging its final hidden states to form a dense sentence-level embedding. Although both shallow (TF-IDF/LSA) and deep (BERT) features were considered during preliminary exploration, only the BERT-based representations were used in downstream training due to their superior ability to capture nuanced semantics in sparse and domain-specific inputs. ~\cite{reimers2019sentence}

All extracted feature vectors were standardized using StandardScaler \cite{pedregosa2011scikit} to facilitate stable convergence in subsequent classifiers. These are later concatenated and standardized to form the final textual representation $ h^T \in \mathbb{R}^{d_T} $, optionally perturbed with Gaussian noise \cite{goodfellow2016deep}.

\subsection*{Image Preprocessing and Feature Extraction}

Device images were sourced from a curated archive of product photographs, encompassing frontal views, schematic diagrams, and contextual usage scenes. To standardize input dimensions and enhance model robustness, all images were initially resized to $256 \times 256$ pixels and then randomly cropped to $224 \times 224$. A range of data augmentation techniques was employed to simulate real-world variability, including horizontal and vertical flipping, affine transformations, grayscale conversion, brightness and contrast jittering, and Gaussian blur. These augmentations help the model generalize better across device images that differ in orientation, lighting, or presentation. Following augmentation, images were normalized using the mean and standard deviation of the ImageNet dataset to align with the input distribution expected by pre-trained convolutional neural networks.

Feature extraction was performed using a fine-tuned EfficientNet-B4 architecture~\cite{tan2019efficientnet}, enhanced with spatial and channel-wise attention modules to emphasize salient regions. The original classification head was removed and replaced with a 64-dimensional bottleneck layer, which served as the final image embedding. The model was trained on the subset of samples with valid image entries using cross-entropy loss \cite{paszke2019pytorch}, the Adam optimizer \cite{kingma2014adam}, and a learning rate scheduler. The resulting image representation is denoted as $h^I \in \mathbb{R}^{d_I}$ and is extracted from the penultimate layer of the network.

\subsection{Multimodal Model Architecture}

\begin{figure}[H]
   \centering
    \includegraphics[width=0.8\textwidth]{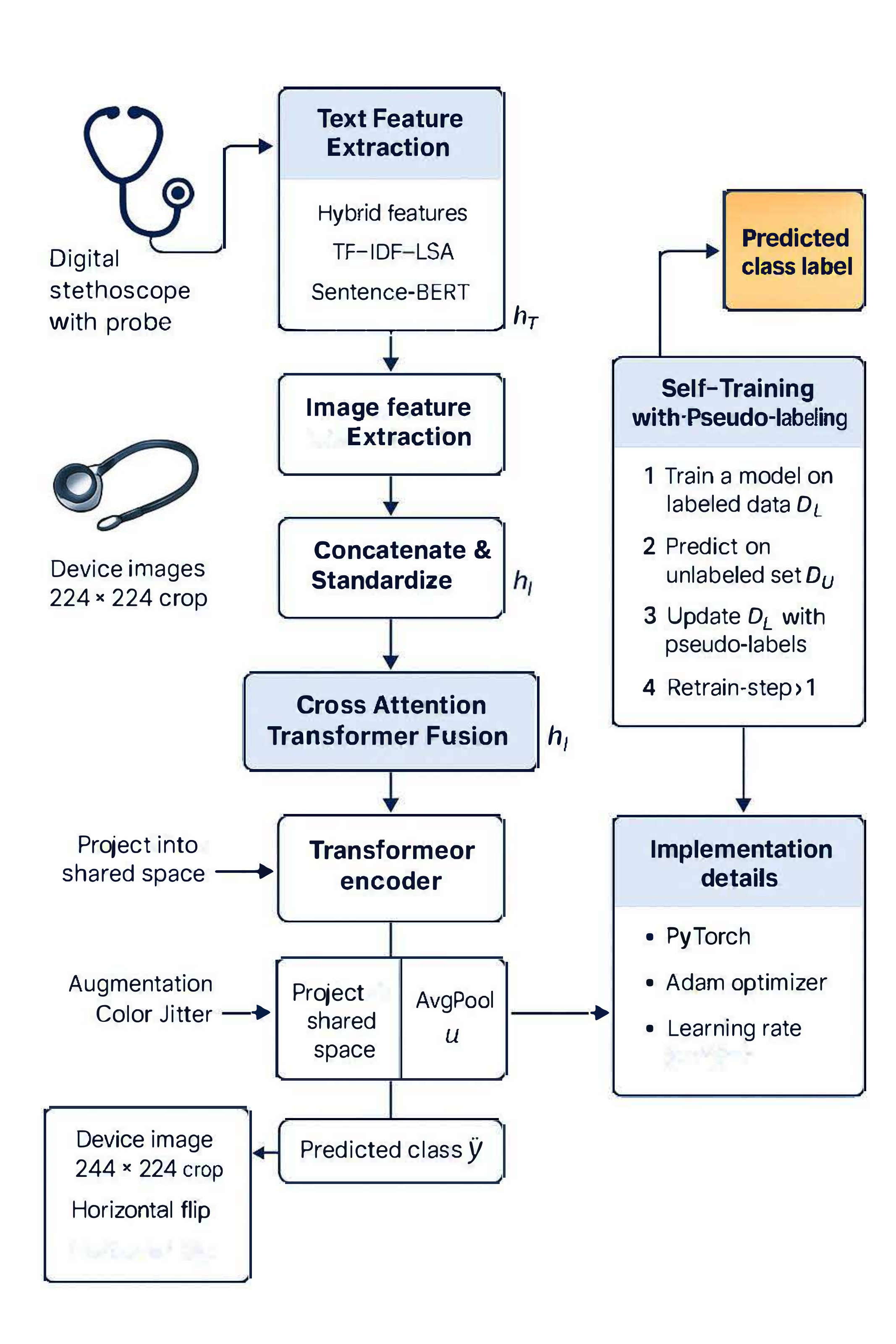}
    \caption{Multimodal Model Training and Evaluation Pipeline}
    \label{fig:per}
\end{figure}

To construct a high-quality aligned multimodal dataset, we ensured that only entries with both valid textual descriptions and matched image files were retained. Sample identifiers were cross-verified using filename patterns against the \texttt{No} column in the structured dataset. Entries failing to meet alignment or preprocessing requirements were discarded. After filtering, the resulting corpus aligned multimodal samples with class-distributed labels.

All features—textual and visual—were independently standardized and optionally reduced in dimensionality via PCA for downstream fusion and classification tasks. This preprocessing pipeline allows us to retain rich semantic and visual signals while mitigating noise and modality imbalance.

To effectively integrate heterogeneous modalities for regulatory classification of medical devices, we propose a transformer-based multimodal architecture that performs joint semantic fusion of textual and visual representations. The model consists of three main components: (1) modality-specific projection layers, (2) a cross-modal Transformer encoder, and (3) a hierarchical feed-forward classifier. This design enables flexible information exchange across modalities while preserving domain-specific signal fidelity.

\subsection*{Modality-Specific Projections}

To map heterogeneous modalities into a common representational space, we apply modality-specific projection networks to the raw text and image embeddings. Sentence-level textual embeddings derived from a pre-trained BERT encoder are first passed through a linear projection layer, followed by batch normalization, GELU activation, and dropout regularization. This process transforms the input into a fixed-length vector $T \in \mathbb{R}^{1 \times d}$, where $d$ is the shared hidden dimension. Similarly, visual features extracted from the EfficientNet-B4 backbone—originally 64-dimensional—are projected through an identical transformation pipeline to yield image embeddings $I \in \mathbb{R}^{1 \times d}$. This parallel design ensures that both modalities are aligned in the same latent space, enabling effective joint reasoning. In our implementation, the hidden dimension $d$ is set to 1024, providing sufficient capacity for capturing complex cross-modal interactions.

\subsection*{Cross-Modal Transformer Encoder}

We concatenate the projected text and image tokens into a sequence $X = [T; I] \in \mathbb{R}^{2 \times d}$ and feed it into a stack of $L$ Transformer encoder layers. Each encoder layer contains multi-head self-attention and position-wise feed-forward sublayers, with residual connections and layer normalization, following the original Transformer design \cite{vaswani2017attention}. Unlike classical text transformers, our encoder performs full cross-modal attention across tokens, allowing contextual information from one modality to modulate the representation of the other.

Formally, given token sequence $X \in \mathbb{R}^{N \times d}$, each self-attention head computes:

\[
\text{Attention}(Q, K, V) = \text{softmax}\left( \frac{QK^\top}{\sqrt{d_k}} \right)V
\]

where $Q$, $K$, and $V$ are linear transformations of $X$, and $d_k = d / h$ is the dimension of each attention head.

The cross-modal attention enables the model to attend to complementary information, improving disambiguation over ambiguous samples.

\subsection*{Multimodal Classifier Head}

The output tokens from the final Transformer layer are flattened and concatenated to form a fused representation $F \in \mathbb{R}^{2d}$. This vector is passed through a feed-forward classification head comprising three fully connected layers with intermediate GELU activations, batch normalization, and dropout. The final output layer maps to $C=3$ classes (Class I, II, III) via a softmax function:

\[
\hat{y} = \text{softmax}(W_3 \cdot \text{GELU}(W_2 \cdot \text{GELU}(W_1 \cdot F)))
\]

where $W_1, W_2, W_3$ are learnable weight matrices. The model is optimized end-to-end using cross-entropy loss between $\hat{y}$ and the ground-truth label $y$.

To handle data scarcity and label imbalance, we employ stratified sampling and oversampling during training. The model is trained for 20 epochs using the Adagrad optimizer (initial learning rate $0.01$, decay $1e$-$4$), with a StepLR scheduler ($\gamma=0.1$ every 5 epochs). Mixed precision training is enabled via PyTorch’s AMP module to accelerate convergence and reduce memory usage.

Our architectural choices are motivated by three key considerations. First, explicit token-level fusion via Transformer layers facilitates context-aware alignment, outperforming implicit early fusion. Second, modality-specific projection ensures representation homogeneity, mitigating modality dominance. Third, the deep feed-forward classifier enables hierarchical feature abstraction, capturing fine-grained distinctions between subtle risk indicators (e.g., ICU-capable vs. home-use variants). This model forms the foundation for our multimodal classification framework and is further enhanced by a self-training strategy described in the following section.

\subsection{Self-Training Strategy}

Self-training is a semi-supervised learning paradigm wherein a model is first trained on labeled data, then used to generate pseudo-labels for unlabeled samples. High-confidence predictions are selected and treated as ground truth in subsequent training rounds. This iterative bootstrapping mechanism allows the model to expand its effective training set and refine its decision boundaries~\cite{yarowsky1995unsupervised, xie2020self}.

To improve the reliability of pseudo-labeling, we adopt an ensemble-based strategy that integrates predictions from multiple base classifiers. Specifically, support vector machines (SVM), logistic regression, and random forest models are trained on the current labeled dataset and then applied to the unlabeled samples. For each sample, we gather the predicted class labels and associated confidence scores from each model. A pseudo-label is retained only if it satisfies two stringent conditions: (i) the average confidence score across all models exceeds a threshold $\tau$ (set to 0.95 in our experiments), and (ii) all models unanimously agree on the predicted class. These dual criteria are designed to ensure the inclusion of only high-quality pseudo-labels, thereby reducing the risk of propagating label noise and mitigating overfitting in subsequent training rounds.

\subsection*{Iterative Augmentation Protocol}

Let $(X_l, y_l)$ denote the initial labeled set and $X_u$ the unlabeled set. The self-training algorithm proceeds as follows:

\begin{enumerate}
    \item Train all models $\{f_i\}$ on $(X_l, y_l)$.
    \item For each $x \in X_u$, compute $\{f_i(x)\}$ and the confidence $\max_i P_i(x)$.
    \item Select a subset $X_s \subset X_u$ that meets both confidence and consistency conditions.
    \item Form $(X_{l}^{+}, y_{l}^{+})$ by adding $(X_s, \hat{y})$ to $(X_l, y_l)$.
    \item Repeat the process for $R$ rounds (we use $R=3$).
\end{enumerate}

In each round, we randomly sample 20\% --- 30\% of the consistent subset to prevent premature overfitting to low-diversity pseudo-labels. The combined dataset is then re-used for model re-training. This strategy is illustrated in Algorithm below.

\begin{algorithm}
\caption{Self-Training with Model Consistency Filtering}
\begin{algorithmic}[1]  
\Require Initial labeled data $(X_l, y_l)$; unlabeled data $X_u$; base models $\{f_i\}$; threshold $\tau$; max rounds $R$
\For{$r = 1$ to $R$}
    \For{each model $f_i$}
        \State Train $f_i$ on $(X_l, y_l)$
    \EndFor
    \State Predict labels and confidence scores on $X_u$
    \State Filter samples where confidence $> \tau$ and all model predictions agree
    \State Select subset $(X_s, \hat{y}_s)$ and augment $X_l \leftarrow X_l \cup X_s$, $y_l \leftarrow y_l \cup \hat{y}_s$
\EndFor
\Ensure Final model trained on expanded $(X_l, y_l)$
\end{algorithmic}
\end{algorithm}

\subsection*{Self-Training with Pseudo-Labeling}

To stabilize training, we use a lightweight StackingClassifier as $f_{\theta^*}$, combining a Random Forest, with a logistic regression meta-learner. This strategy provides more robust pseudo-labels during early training rounds. The student model is iteratively updated using newly added pseudo-labeled data, and the student is promoted to the new teacher in the next round.

To ensure the validity of our self-training procedure, we strictly separated the training and evaluation datasets. The self-training loop was conducted exclusively on the labeled training set, where pseudo-labels were generated using the Stacking model and applied only to unlabeled training samples. A confidence threshold of 0.9 was used to select high-confidence pseudo-labeled samples, which were then added to the training set for the next round. No samples from the held-out test set were used in any form—neither for pseudo-labeling, model fitting, nor evaluation—thus maintaining a clear separation between training and evaluation phases. During evaluation, we employed 5-fold stratified cross-validation solely on the (labeled + pseudo-labeled) training data to assess model performance. The reported metrics (Accuracy, F1-score, Precision, Recall, AUROC) are averaged across the folds and computed strictly on validation subsets that were not seen during training. This setup adheres to standard semi-supervised learning protocols and avoids any form of data leakage.

We adopt self-training~\cite{xie2020self} with the following procedure:
\begin{enumerate}
    \item Train model on labeled data $ \mathcal{D}_L $
    \item Predict on unlabeled set $\mathcal{D}_U $
    \item Select samples with confidence $ \max(p(y)) > \tau = 0.9 $
    \item Add them to $ \mathcal{D}_L $ with predicted labels
    \item Retrain for $R=3$ rounds
\end{enumerate}

Inspired by Noisy Student Training \cite{xie2020self}, we design a self-training loop that leverages multimodal features and stacking-based pseudo-labeling. Let the labeled dataset be defined as $\mathcal{D}_L = \{(\mathbf{x}_i^{(t)}, \mathbf{x}_i^{(v)}, y_i)\}_{i=1}^{n}$, where $\mathbf{x}_i^{(t)}$ and $\mathbf{x}_i^{(v)}$ denote textual and visual features respectively. Let $\mathcal{D}_U = \{(\tilde{\mathbf{x}}_j^{(t)}, \tilde{\mathbf{x}}_j^{(v)})\}_{j=1}^{m}$ be the unlabeled dataset.

At each iteration of self-training, a teacher model $f_{\theta^*}$ is trained on the labeled data. Pseudo-labels are generated for the unlabeled data using:

\begin{equation}
\tilde{y}_j = \arg\max_k f_{\theta^*}^{(k)}(\tilde{\mathbf{x}}_j^{(t)}, \tilde{\mathbf{x}}_j^{(v)}), \quad \text{subject to } \max_k f_{\theta^*}^{(k)}(\cdot) > \tau
\label{eq:pseudo-label}
\end{equation}

where $f_{\theta^*}^{(k)}(\cdot)$ denotes the predicted probability for class $k$, and $\tau$ is a fixed confidence threshold (e.g., $\tau=0.9$). Only samples with high-confidence predictions are included in the pseudo-labeled dataset.

The student model $f_\theta$ is then trained to minimize a combined loss on both labeled and pseudo-labeled data:

\begin{equation}
\mathcal{L}_{\text{total}}(\theta) = \frac{1}{n} \sum_{i=1}^{n} \ell\left(y_i, f_{\theta}(\text{Aug}(\mathbf{x}_i^{(t)}, \mathbf{x}_i^{(v)}))\right) + \lambda \cdot \frac{1}{m} \sum_{j=1}^{m} \ell\left(\tilde{y}_j, f_{\theta}(\text{Aug}(\tilde{\mathbf{x}}_j^{(t)}, \tilde{\mathbf{x}}_j^{(v)}))\right)
\label{eq:total-loss}
\end{equation}

Here, $\ell(\cdot)$ denotes the cross-entropy loss, and $\text{Aug}(\cdot)$ represents optional input perturbation such as dropout or feature masking. We set $\lambda=1$ in our experiments.

\subsection*{Implementation Details}

All models are implemented in PyTorch. Textual preprocessing is conducted using Scikit-learn for TF-IDF and LSA extraction, and the SentenceTransformers library for Sentence-BERT embeddings. Image feature extraction utilizes EfficientNet-B4 from the \texttt{efficientnet\_pytorch} package.

The Transformer encoder used in multimodal fusion consists of $L=4$ layers, each with $16$ attention heads, a hidden dimension of $1024$, a feedforward expansion factor of $4\times$, and dropout rates of $0.2$ in the attention and feedforward blocks. Cross-modal representations are first projected to a $d=1024$ latent space before fusion.

Training is performed using the Adam optimizer with an initial learning rate of $3\times10^{-4}$, a batch size of $32$, and early stopping with a patience of $10$ epochs, up to a maximum of $50$ epochs. For self-training, an ensemble comprising Support Vector Machine (SVM), Logistic Regression, and Random Forest classifiers is employed. Pseudo-labels are accepted if the average confidence across models exceeds a threshold of $0.95$ and predictions are consistent.

Data augmentation for images includes random resized cropping (to $224\times224$), horizontal and vertical flipping, random rotation (up to $45^\circ$), color jittering, random grayscaling with probability $0.2$, Gaussian blurring, and random perspective distortion.

%% file: sections/experiments.tex
\section{Experiments and Results}
\label{sec:experiments}

To evaluate the effectiveness of our proposed multimodal classification framework, we conduct experiments comparing text-only, image-only, and fused modalities under various training strategies. We report quantitative results on classification accuracy, F1-score, precision, recall, and AUROC across multiple model types. Additional ablation studies assess the contribution of each modality and the impact of self-training. We adopt a five-fold stratified cross-validation protocol to ensure statistical robustness across imbalanced class distributions. Models are evaluated on the held-out folds, and reported metrics reflect the mean performance across all splits. For each fold, we fix the same training/validation partition across all methods to enable fair comparison.

We compare against three strong classical classifiers:
\begin{itemize}
    \item \textbf{Support Vector Machine (SVM)} with RBF kernel;
    \item \textbf{Logistic Regression (LR)} with L2 regularization;
    \item \textbf{Random Forest (RF)} with 300 estimators and max depth 15.
\end{itemize}

Each model is tested under three settings: (1) text-only features, (2) image-only features, and (3) early-fusion of text and image features. For the proposed model, we evaluate both with and without self-training enhancement.

\begin{figure}[t]
    \centering
    \includegraphics[scale=0.5]{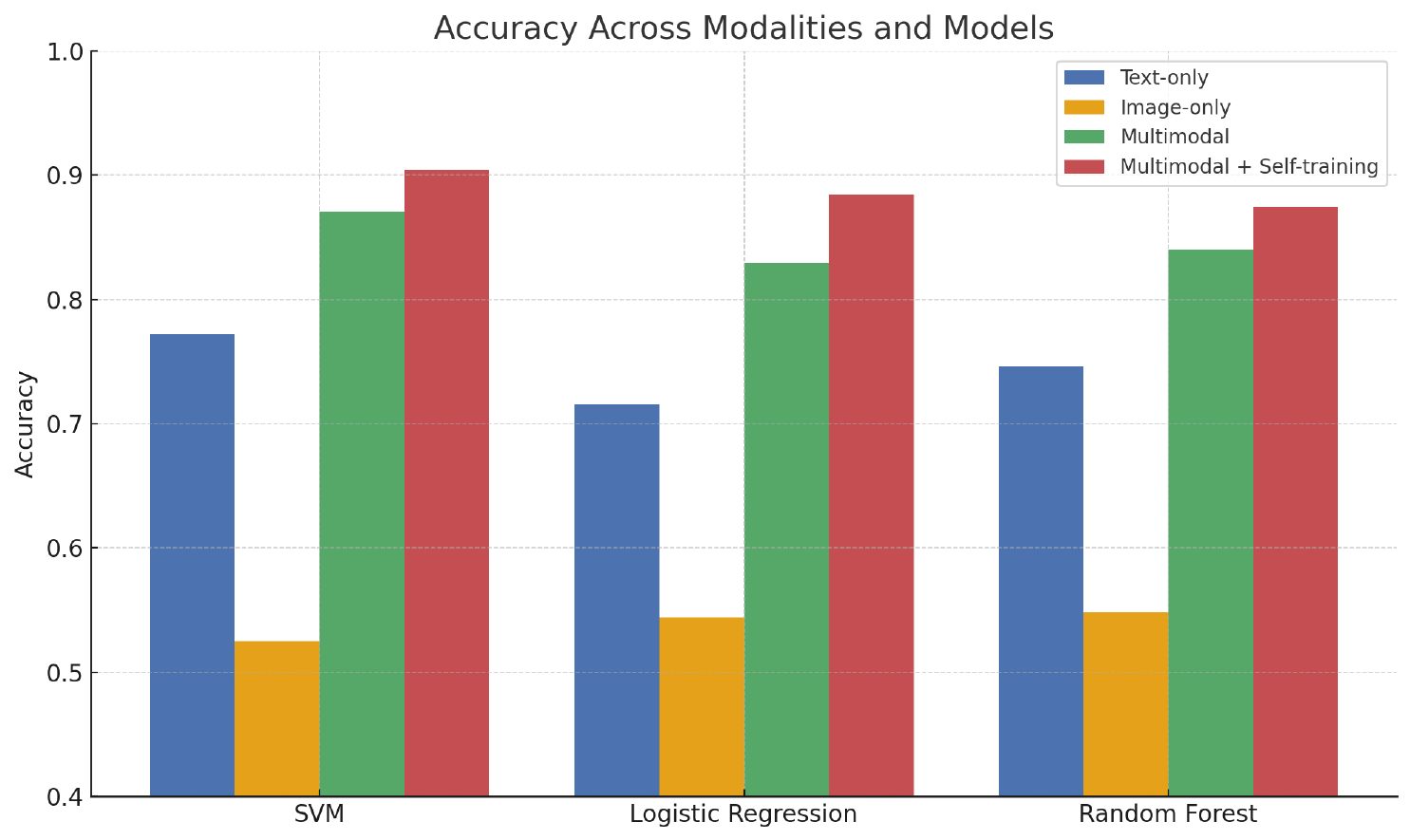}
    \caption{Predication Accuracy Rate Across Different Models}
    \label{fig:performance}
\end{figure}

Table~\ref{tab:text_image}, Table \ref{tab:multimodal} and Figure \ref{fig:performance} summarizes the performance across different modalities and classifiers. We observe several consistent trends across experimental conditions. Text-only models demonstrate reasonable classification performance, primarily due to the presence of domain-specific linguistic cues in the device descriptions. In contrast, image-only models exhibit less stable behavior, likely attributable to the inherent variability in image resolution, framing, and visual noise across the dataset. Notably, the multimodal fusion approach outperforms all unimodal baselines across classifiers, confirming the value of joint modeling in capturing complementary semantic signals. Moreover, the introduction of self-training yields additional gains, improving classification accuracy, with particularly notable benefits in underrepresented classes where labeled examples are scarce.

\begin{table}[htbp]
\centering
\small
\caption{Performance Comparison: Text-only and Image-only Models}
\label{tab:text_image}
\begin{tabular}{llrrrrr}
\toprule
\textbf{Model} & \textbf{Setting} & \textbf{Accuracy} & \textbf{F1} & \textbf{Precision} & \textbf{Recall} & \textbf{AUROC} \\
\midrule
SVM & Text-only  & 0.7720 & 0.7706 & 0.7719 & 0.7703 & -- \\
Logistic Regression & Text-only & 0.7155 & 0.7140 & 0.7153 & 0.7141 & -- \\
Random Forest & Text-only & 0.7458 & 0.7446 & 0.7457 & 0.7445 & -- \\
SVM & Image-only  & 0.5252 & 0.5130 & 0.5154 & 0.5151 & -- \\
Logistic Regression & Image-only & 0.5438 & 0.5288 & 0.5331 & 0.5315 & -- \\
Random Forest & Image-only & 0.5481 & 0.4926 & 0.5278 & 0.5198 & -- \\
\bottomrule
\end{tabular}
\end{table}

\begin{table}[htbp]
\centering
\small
\caption{Performance Comparison: Multimodal and Multimodal + Self-training Models}
\label{tab:multimodal}
\begin{tabular}{llrrrrr}
\toprule
\textbf{Model} & \textbf{Setting} & \textbf{Accuracy} & \textbf{F1} & \textbf{Precision} & \textbf{Recall} & \textbf{AUROC} \\
\midrule
SVM & Multimodal & 0.8711 & 0.8710 & 0.8712 & 0.8710 & 0.9646 \\
Logistic Regression & Multimodal & 0.8294 & 0.8284 & 0.8284 & 0.8294 & 0.9287 \\
Random Forest & Multimodal & 0.8400 & 0.8396 & 0.8419 & 0.8399 & 0.9464 \\
SVM & Multimodal (Self training) & 0.9038 & 0.8854 & 0.8877 & 0.8857 & 0.9788 \\
Logistic Regression & Multimodal (Self training) & 0.8841 & 0.8624 & 0.8652 & 0.8624 & 0.9535 \\
Random Forest & Multimodal (Self training) & 0.8743 & 0.8503 & 0.8533 & 0.8508 & 0.9669 \\
\bottomrule
\end{tabular}
\end{table}

\subsection*{Ablation Study: Impact of Visual Modality}

To assess the importance of visual information, we conduct an ablation experiment in which image features are replaced with Gaussian noise of matching dimensions. The resulting model (Text + Image Noise) demonstrates a significant drop in performance, confirming that the visual modality contributes meaningfully to prediction (Table~\ref{tab:ablation}).

\begin{table}[htbp]
\centering
\small
\caption{Ablation Test Results (Image Features Replaced with Noise)}
\label{tab:ablation}
\begin{tabular}{lccccc}
\toprule
\textbf{Model} & \textbf{Accuracy} & \textbf{F1 Score} & \textbf{Precision} & \textbf{Recall} & \textbf{AUROC} \\
\midrule
SVM & 0.5272 & 0.2301 & 0.8424 & 0.3333 & 0.5265 \\
Logistic Regression & 0.5272 & 0.2301 & 0.8424 & 0.3333 & 0.6076 \\
Random Forest & 0.5449 & 0.3624 & 0.6287 & 0.3641 & 0.5948 \\
\bottomrule
\end{tabular}
\end{table}

Our results confirm the effectiveness of transformer-based multimodal fusion and self-training for medical device risk classification. Multimodal models substantially outperform unimodal baselines, and self-training yields further improvements by leveraging unlabeled examples with high-confidence pseudo-labels. The combination of cross-modal attention and semi-supervised learning provides a scalable solution for real-world regulatory data scarcity.

%% file: sections/conclusion.tex
\section{Discussion}

This study proposes a transformer-based multimodal framework for medical device risk classification that jointly integrates textual descriptions and visual information extracted from regulatory documents. By leveraging cross-modal attention mechanisms and a self-training enhancement strategy, the approach captures fine-grained semantic dependencies between modalities and improves generalization under limited supervision. Extensive experiments across multiple classifiers and fusion strategies demonstrate that the proposed architecture consistently outperforms unimodal and traditional fusion baselines by a substantial margin.

Beyond technical performance, the practical implications of the findings merit deeper consideration. The proposed framework could serve as a valuable assistive tool for regulatory professionals by automating the preliminary classification of medical devices, thereby alleviating reviewer workload and improving consistency in market approval processes. Particularly for high-volume, lower-risk categories (e.g., Class I and II devices), the model offers potential to streamline initial screening procedures. In manufacturing contexts, real-time integration into factory-level inspection systems could enable automated verification of device risk classes, reducing mislabeling incidents and enhancing compliance audit efficiency. Furthermore, the modality-agnostic feature representations learned from jointly modeling text and images offer resilience to documentation variations across jurisdictions, supporting broader efforts toward cross-border regulatory harmonization (e.g., UDI and GMDN systems).

Nevertheless, several limitations constrain the scope and generalizability of the results. The dataset used in this study, while carefully curated, remains relatively modest in size and is exclusively derived from the Chinese NMPA regulatory database. This single-source origin inherently limits the diversity of device categories, regulatory language, and documentation formats captured, and caution is warranted when extrapolating the findings to other regulatory environments. Expanding future studies to include data from multiple agencies, such as the FDA or EMA, would provide a stronger basis for generalization.

Moreover, although the model effectively fuses textual and visual modalities, it does not yet incorporate structured metadata fields—such as intended clinical use, manufacturer identity, or device composition—that are often central to regulatory assessments. Including such structured information could further enhance classification fidelity, particularly for borderline or novel device types. The self-training mechanism, while successful in leveraging high-confidence pseudo-labels to augment the training set, rests on the assumption that confidence correlates with correctness. In domains with imbalanced class distributions or evolving device categories, this assumption may not always hold, introducing risks of confirmation bias during label propagation.

Overall, this work lays a foundation for developing scalable, interpretable, and domain-sensitive AI systems to support regulatory science. Future research could focus on integrating structured metadata, strengthening model explainability (e.g., through SHAP-based interpretation of classification decisions), and validating model performance in collaboration with regulatory professionals under real-world operational conditions. Bridging technical advancements with the realities of regulatory practice will be critical for translating AI innovation into tangible benefits for medical device oversight.

\section{Conclusion and Future Work}

This work presents, to the best of our knowledge, the first application of a Transformer-based multimodal fusion model with self-training for medical device risk classification. Our framework outperforms both unimodal baselines and early-fusion strategies, demonstrating the effectiveness of multimodal learning in the context of regulatory informatics. The results highlight that visual features substantially enhance disambiguation of device risk levels, and that high-confidence pseudo-labeling can improve generalization under limited supervision. By integrating advances in vision-language modeling and semi-supervised learning, this framework establishes a scalable foundation for future research at the intersection of regulatory AI and medical informatics.

Several promising directions exist for extending this work to achieve greater robustness and applicability in real-world regulatory settings. An important avenue involves adapting the framework for multilingual and cross-jurisdictional deployment. Medical device classification protocols vary considerably across regions, including systems such as the FDA (United States), NMPA (China), and MDR (European Union). Developing a unified model capable of accommodating regulatory heterogeneity—and processing documentation in multiple languages—would provide a critical step toward globally interoperable AI decision support systems.

Incorporating structured metadata alongside unstructured modalities represents another natural extension. Clinical specialty indicators, standardized product codes, and manufacturer identifiers offer valuable domain-specific context that is often underutilized. Jointly modeling these structured signals with visual and textual embeddings could further enrich the model's representational capacity, particularly in edge cases where modality cues are ambiguous or inconsistent. Given the dynamic nature of medical device innovation, strategies for enhancing model generalization—such as zero-shot or few-shot learning—also warrant exploration. Pretraining with contrastive objectives or instruction tuning, as demonstrated in large-scale vision-language models, could enable the system to recognize emerging device types with minimal supervision, addressing a key challenge in rapidly evolving regulatory landscapes.

Future work could incorporate explanation mechanisms, such as attention-based visualizations or SHAP-based attribution maps, to illuminate the rationale behind risk predictions. Such capabilities are critical not only for fostering user trust but also for supporting regulatory auditing, documentation, and accountability. These directions outline a path toward building more scalable, adaptive, and trustworthy AI systems for medical device classification, advancing the integration of machine learning into regulatory science and clinical technology governance.

\section*{DATA AVAILABILITY}
The scripts used to carry out the analysis and the resultant data that support the findings in this study are available on: https://github.com/RegAItool/Multimodal-regulatory